\documentclass{article}

\newenvironment{enumerate*}%
  {\begin{enumerate}%
    \setlength{\itemsep}{.2pt}%
    \setlength{\parskip}{.2pt}%
    \setlength{\topsep}{.5pt}}%
  {\end{enumerate}}

\newenvironment{itemize*}%
  {\begin{itemize}%
    \setlength{\leftmargin}{-.2in}%
    \setlength{\itemsep}{.2pt}%
    \setlength{\parskip}{.2pt}%
    \setlength{\topsep}{.5pt}}%
  {\end{itemize}}

\usepackage{times}
\usepackage{graphicx} 
\usepackage{subcaption} 

\usepackage{amsmath}
\usepackage{amssymb}

\usepackage{natbib}

\usepackage{algorithm}
\usepackage{algorithmic}

\usepackage{multirow}

\usepackage{hyperref}

\usepackage[juststyle]{icml2017} 

\icmltitlerunning{Forced to Learn: Discovering Disentangled Representations Without Exhaustive Labels}

\begin{document} 

\twocolumn[
\icmltitle{Forced to Learn: Discovering Disentangled Representations Without Exhaustive Labels}



\icmlsetsymbol{equal}{*}

\begin{icmlauthorlist}

\icmlauthor{Alexey Romanov}{uml}
\icmlauthor{Anna Rumshisky}{uml}
\end{icmlauthorlist}

\icmlaffiliation{uml}{Department of Computer Science, University of Massachusetts Lowell, Lowell, Massachusetts, USA}

\icmlcorrespondingauthor{Alexey Romanov}{aromanov@cs.uml.edu}

\icmlkeywords{deep learning, representation, clustering, loss, neural networks, kmeans}

\vskip 0.3in
]



\printAffiliationsAndNotice{} 

\begin{abstract} 
Learning a better representation with neural networks is a challenging problem, which was tackled extensively from different prospectives in the past few years. In this work, we focus on learning a representation that could be used for a clustering task and introduce two novel loss components that substantially improve the quality of produced clusters, are simple to apply to an arbitrary model and cost function, and do not require a complicated training procedure. We evaluate them on two most common types of models, Recurrent Neural Networks and Convolutional Neural Networks, showing that the approach we propose consistently improves the quality of KMeans clustering in terms of Adjusted Mutual Information score and outperforms previously proposed methods.

\end{abstract}

\section{Introduction}
\label{sec:introduction}

In the past few years, a substantial amount of work has been dedicated to learning a better representation of the input data that can be either used in downstream tasks, such as KMeans clustering, or to improve generalizability or perfromance of the model. In general, these works can be divided into two categories:
\begin{enumerate*}
\item[(1)] approaches that require a complicated training procedure;
\item[(2)] approaches that introduce a new loss component that can be easily applied to an arbitrary cost function;
\end{enumerate*}
For example, approaches by \citet{liao2016learning} and \citet{xie2016unsupervised} can be assigned to the first category, as they propose to iteratively refine the clusters during the training. In contrast, approaches by \citet{cogswell2015reducing} and Cheung et~al.~\citet{cheung2014discovering} introduce new loss components that can be added to the cost function while training the model with a standard gradient descent algorithm. 
Our work belongs to the second category and focuses on a challenging problem of learning disentangled representations while having access to labels that do not fully reflect the underlying partitioning of the data, but still separate it into distinguishable groups. 

\begin{figure}[ht]
\vskip 0.2in
\begin{center}
\centering
    \includegraphics[width=1\columnwidth, clip, trim=0.7cm 0.7cm 0.7cm 0.7cm]{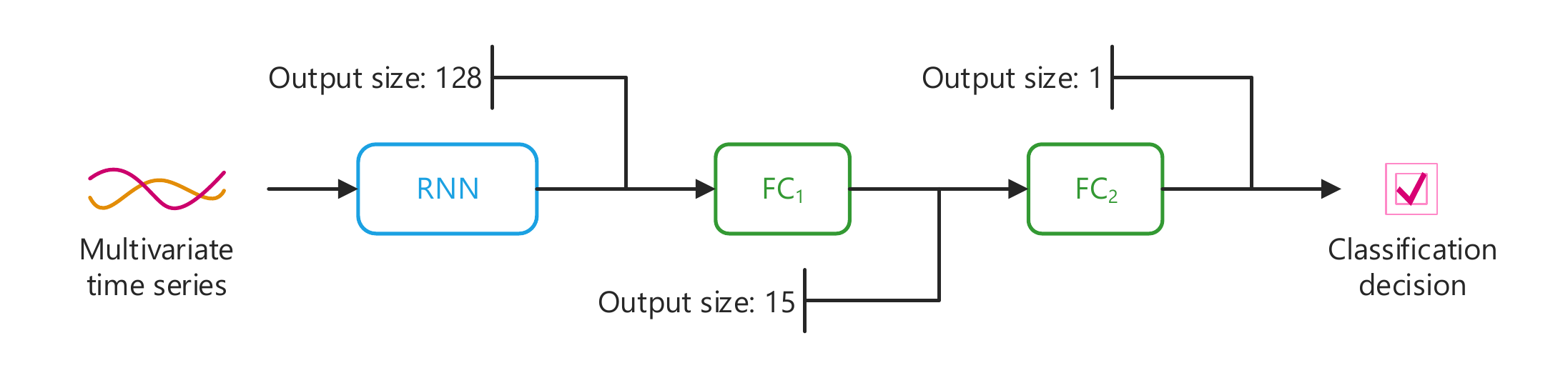}
\caption{An RNN model with two fully-connected layers for binary classification of time series}
\label{fig:rnn_model}
\end{center}
\vskip -0.2in
\end{figure}

\begin{figure*}
\vskip 0.2in
\begin{center}
\begin{subfigure}{1\columnwidth}
\centering
  \includegraphics[width=0.8\columnwidth]{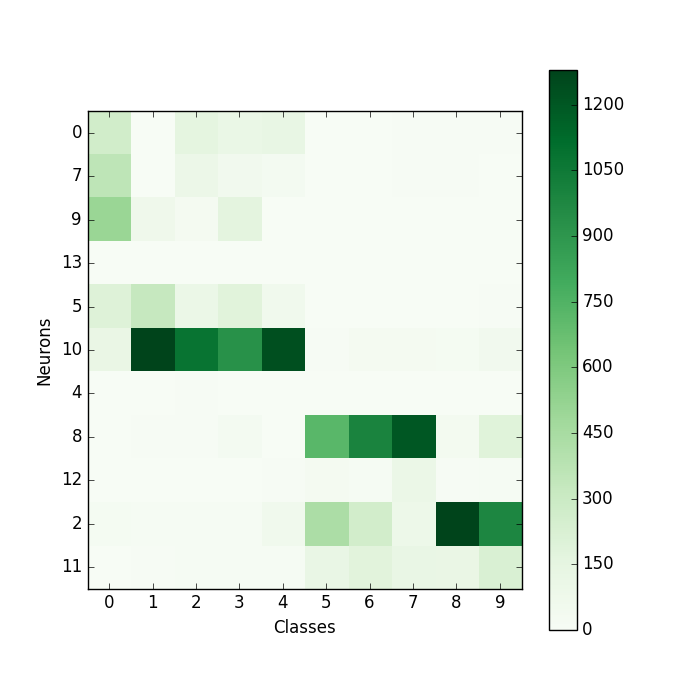}
  \caption{Without the proposed loss component}
  \label{fig:neurons_without_loss}
\end{subfigure}
\begin{subfigure}{1\columnwidth}
\centering
  \includegraphics[width=0.8\columnwidth]{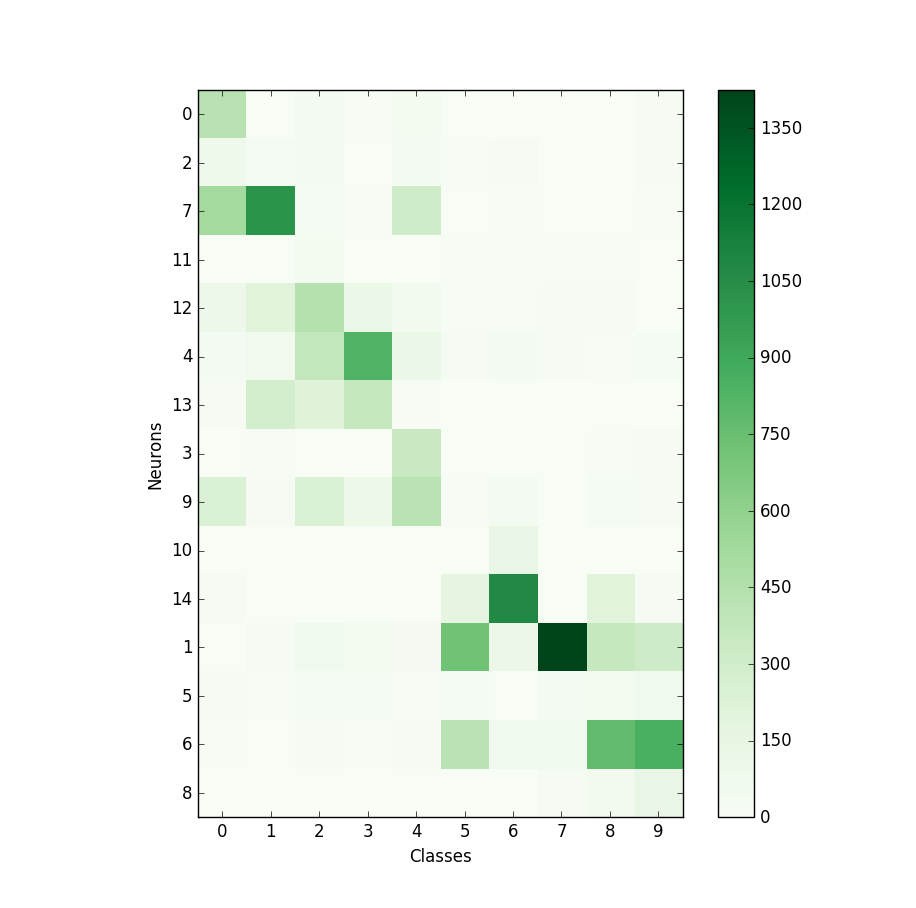}
  \caption{With the proposed loss component $\mathcal{L}_{\text{single}}$}
  \label{fig:neurons_with_loss}
\end{subfigure}
\caption{Number of samples for which the neurons on the $y$ axis were active the most in a binary classification task on MNIST strokes sequences dataset. The classes 0-4 have the label 0, and the classes 5-9 have the label 1. See the~\autoref{sec:experiments_mnist} for details.}
\label{fig:neurons}
\end{center}
\vskip -0.2in
\end{figure*}

For example, consider a case of predicting in-hospital mortality using multivariate physiological time series. This is a binary classification problem which can be solved using an Recurrent Neural Network model such as the one depicted on \autoref{fig:rnn_model}. During a regular training procedure with a sigmoid cross-entropy loss, the model tends to learn the  weights that lead to a strong activation of one of the neurons in the penultimate layer ($\text{FC}_1$) for the instances that belong to the positive class and a strong activation of another neuron for the instances that belong to the negative class, whereas all other neurons tend to be not active for both classes (see the~\autoref{fig:neurons_without_loss}). 
However, we would like to separate patients into more than two groups by applying a clustering algorithm to the learned representations of the patients. Thus, we would need the model to learn a disentangled representation that can not only differentiate between the patients with different outcomes, but also between the patients with the same outcome using latent characteristics of the time series (see the~\autoref{fig:neurons_with_loss}).

In order to force the network to learn such disentangled representations, we propose two novel loss components that can be applied to an arbitrary cost function. Although it can be used in any type of model, including autoencoders, this paper focuses on a task of learning disentangled representations during the binary classification problem.

\section{Related Work}
\label{sec:related_work}
In the past few years we witnessed an astonishing success of neural networks. Starting with ILSVRC in 2012~\cite{ILSVRC15}, where a deep convolutional neural network model won the challenge with a shocking gap~\cite{krizhevsky2012imagenet}, neural networks has achieved remarkable success in nearly every classification task.
For a long time, it was unclear why deep neural networks, and convolutional neural networks (CNN) in particular, work so well and what is happening inside this ``black box'', until the work of~\citet{zeiler2014visualizing}, which showed that it is possible to visualize which features and representations specifically a network learns during the training, which answered this question for CNNs, but not for Recurrent Neural Networks (RNN). More recently, there have been a few works that sought to perform a similar analysis to RNNs. In particular, the work of~\citet{karpathy2015visualizing} showed what is happening inside an RNN cell during the inference and the areas of responsibility of neurons inside the cell. After that, \citet{li2016visualizing} plotted the representations learned by an RNN trained on a task of sentiment classification and showed that the network is able to learn local compositionality, embedding the negated expressions (such as ``not good'', ``not nice'') into the space near words with a negative polarity (such as ``bad'').

More relevant to the goal of this paper, \citet{cheung2014discovering} proposed a cross-covariance penalty (XCov) to force the network to produce representations with disentangled factors. The proposed penalty is, essentially, cross-covariance between the predicted labels and the activations of samples in a batch. Their experiments showed that the network can produce a representation, with components that are responsible to different characteristics of the input data. For example, in case of the MNIST dataset, there was a class-invariant factor that was responsible for the style of the digit, and in case of the Toronto Faces Dataset~\cite{susskind2010toronto}, there was a factor responsible for the subject's identity.
Similarly, but with a different goal in mind, \citet{cogswell2015reducing} proposed a new regularizer (DeCov) which minimizes cross-covariance of hidden activations, leading to non-redundant representations and, consequently, less overfitting and better generalization. DeCov loss is trying to minimize the Frobenius norm of the covariance matrix between all pairs of activations in the given layer. The authors' experiments showed that the proposed loss significantly reduced overfitting and led to a better classification performance on a variety of datasets.

\begin{figure*}[th!]
\vskip 0.2in
\begin{center}
\begin{subfigure}{1\columnwidth}
\centering
  \includegraphics[width=0.8\columnwidth]{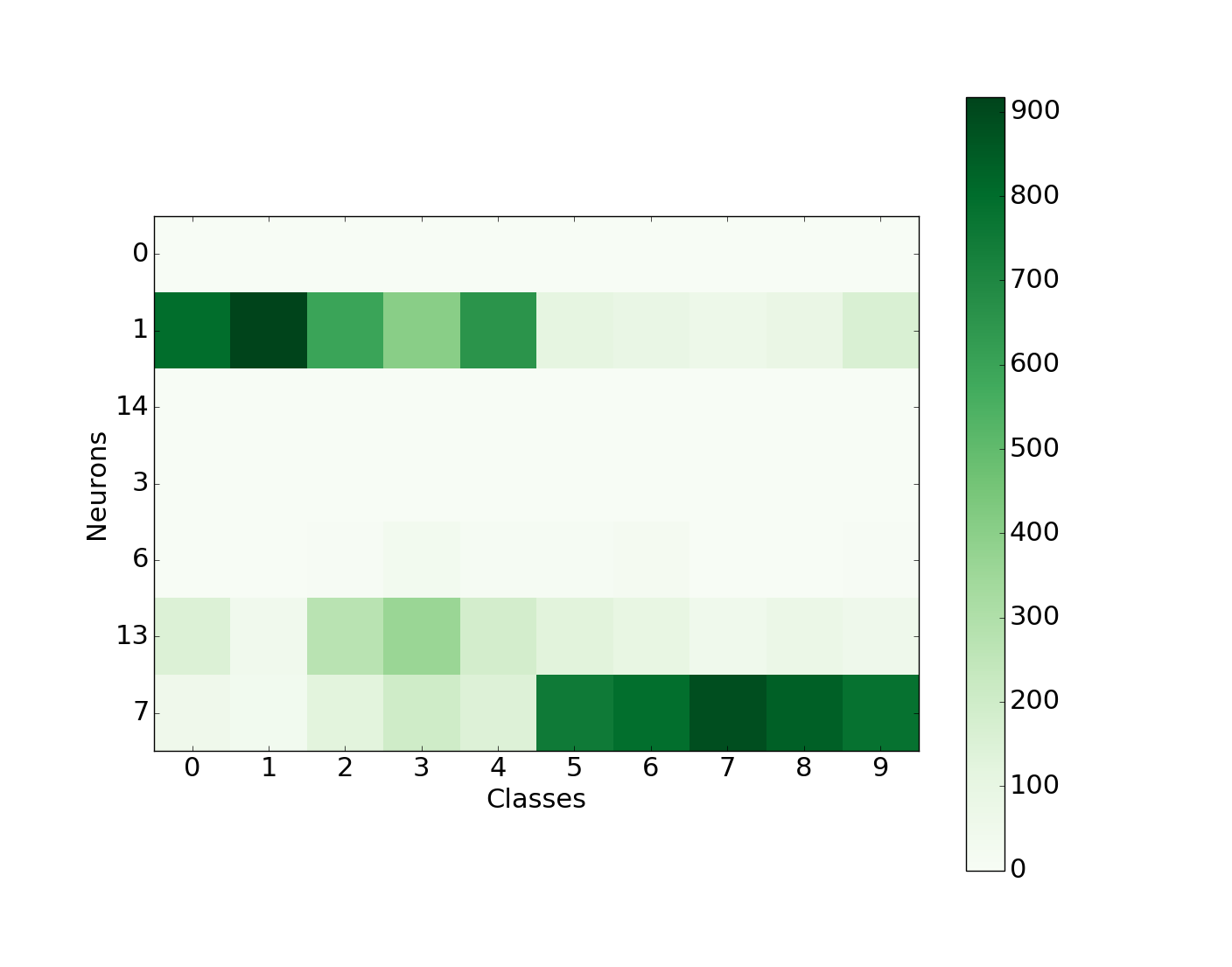}
  \caption{Without the proposed loss component}
  \label{fig:neurons_cifar10_without_loss}
\end{subfigure}
\begin{subfigure}{1\columnwidth}
\centering
  \includegraphics[width=0.8\columnwidth]{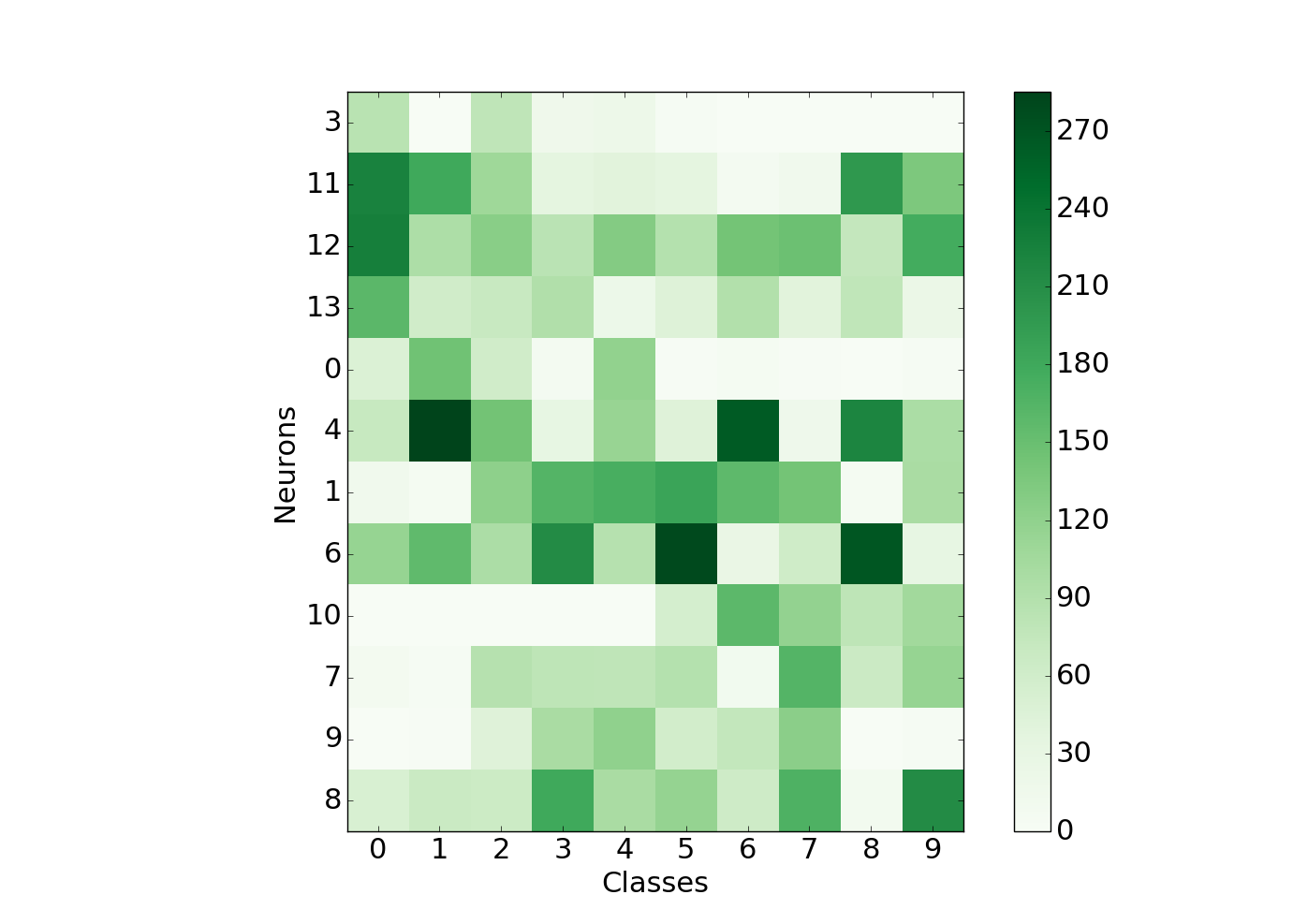}
  \caption{With the proposed loss component $\mathcal{L}_{\text{multi}}$}
  \label{fig:neurons_cifar10_with_loss}
\end{subfigure}
\caption{Number of samples for which the neurons on the $y$ axis were active the most in a binary classification task on the CIFAR-10 dataset. See the~\autoref{sec:experiments_cifar10} for details.}
\label{fig:neurons_cifar10}
\end{center}
\vskip -0.2in
\end{figure*}

While the aforementioned approaches do not require a complicated training procedure, there were also works that proposed more convoluted algorithms in order to obtain better representations that could be used in downstream tasks, such as clusterization. \citet{liao2016learning} proposed a method to learn parsimonious representations. Essentially, the proposed algorithm iteratively calculates cluster centroids, which are updated every $M$ iterations and used in the cost function. The authors' experiments showed that such algorithm leads to a better generalization and a higher test performance of the model in case of supervised learning, as well as unsupervised and even zero-shot learning. Similarly, \citet{xie2016unsupervised} proposed an iterative algorithm that first calculates soft cluster assignments, then updates the weights of the network and cluster centroids. This process is repeated until convergence. In contrast to~\citet{liao2016learning}, the authors specifically focused on the task of learning better representations for clustering, and showed that the proposed algorithm gives a significant improvement in clustering accuracy.

In contrast to the last two methods, our proposed loss components do not require such complicated training procedures and can be conveniently used in any cost function in a straightforward manner.

\section{The proposed method}
\label{sec:proposed_method}

Inspired by the work of~\citet{cheung2014discovering} and~\citet{cogswell2015reducing}, we propose two novel loss components which despite their simplicity, significantly improve the quality of the clustering over the representation produced by the model. The first loss component $\mathcal{L}_{\text{single}}$ works on a single layer and does not affect the other layers in the network, which may be a desirable behaviour in some cases. The second loss component $\mathcal{L}_{\text{multi}}$ affects the entire network behind the target layer and forces it to produce disentangled representations in more complex and deep networks in which the first loss may not give the desired improvements.

\subsection{Single layer loss}
\label{sec:single_layer_loss}

Consider the model on the~\autoref{fig:rnn_model}. The layer $\text{FC}_2$ has the output size of 1 and produces a binary classification decision. The output of the layer $\text{FC}_1$ is used to perform KMeans clustering. Recall from the example in the introduction that we want to force the model to produce divergent representations for the samples that belong to the same class, but are in fact substantively different from each other. One way to do it would be to force the \textit{rows} of the weight matrix $W_{\text{FC}_1}$ of the $\text{FC}_1$ layer be different from each other, leading to different patterns of activations in the output of the $\text{FC}_1$ layer.

Formally, it can be expressed as follows:
\begin{equation}
    \label{eq:loss_single_sum}
    \mathcal{L}_{\text{single}} = \sum_{i=1}^{k}\sum_{j=i+1}^{k} f_l(d_i,d_j) + f_l(d_j,d_i)
\end{equation}
where $d_k$ are normalized weights of the row $k$ of the weights matrix $W$ of the given layer: 
\begin{equation}
    d_k = \text{softmax}(W[k])
\end{equation}
and $f_l(d_i,d_j)$ is a component of the loss between the rows $i$ and $j$:
\begin{equation}
    \label{eq:f_l}
    f_l(x_i,x_j) = \max(0, m - D_{\text{KL}}(x_i || x_j))
\end{equation}
where $m$ is a hyperparameter that defines the desired margin of the loss component and $D_{\text{KL}}(d_i || d_j)$ is the Kullback-Leibler divergence between the probability distributions $d_i$ and $d_j$.

\begin{figure*}[ht]
\vskip 0.2in
\begin{center}
\centering
    \includegraphics[width=1\linewidth, clip, trim=0.7cm 0.7cm 0.7cm 0.7cm]{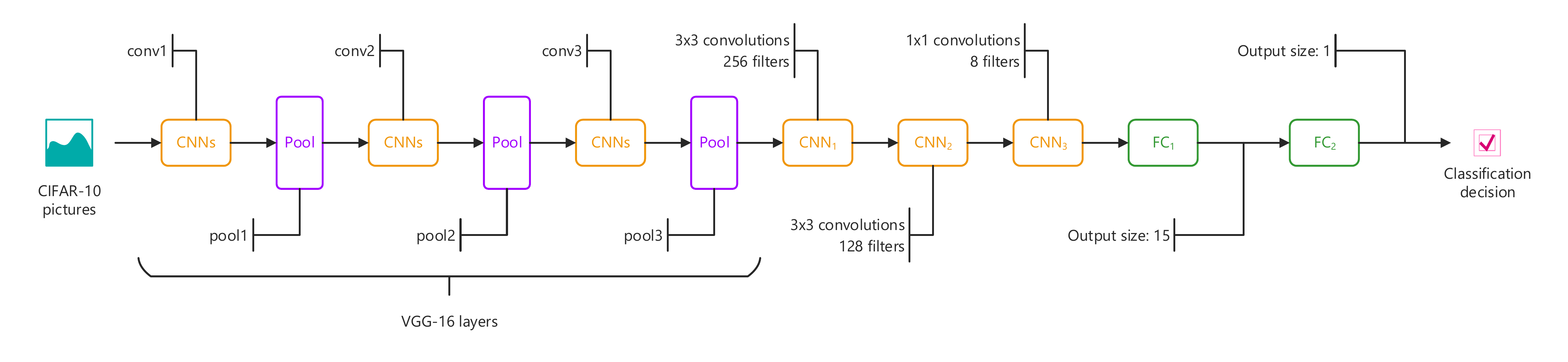}
\caption{A CNN model used in the CIFAR-10 experiments}
\label{fig:cnn_model}
\end{center}
\vskip -0.2in
\end{figure*}

\subsection{Multilayer loss}
\label{sec:multilayer_loss}

Note that the loss component $\mathcal{L}_{\text{single}}$ affects only the weights of the specific layer, as it operates not on the outputs of the layer but directly on its weights, similar to, for example, $\ell_2$ regularization. Therefore, this loss component may help to learn a better representation only if the input to the target layer still contains the information about latent characteristics of the input data. This might be the case in simple shallow networks, but in case of very deep complex networks the input data is non-linearly transformed so many times that only the information that is needed for binary classification left, and all the remaining latent characteristics of the input data were lost as not important for binary classification (see the~\autoref{fig:neurons_cifar10_without_loss}). Indeed, as we can see from the experiments in Section~\ref{sec:experiments}, the loss component described above substantially improves the quality of the clustering in a simple baseline case. However, in case of a more complex model, this improvement is much less impressive. Therefore, we are also proposing a loss component that can influence not only one specific layer, but all layers before it, in order to force the network to produce a better representation. 

Recall again that we want to force the model to produce disentangled representations of the input data. Namely, that these representations should be sufficiently different from each other even if two samples have the same label.
We propose the following loss component in order to produce such properties:
\begin{equation}
    \label{eq:loss_multi_sum}
    \mathcal{L}_{\text{multi}} = \frac{1}{N_s^2} \sum_{i=1}^{N}\sum_{j=1}^{N} 
    \begin{cases}
        f_l(h_i^s,h_j^s) + f_l(h_j^s,h_i^s) & y_i = y_j \\
        0 & y_i \neq y_j
    \end{cases}
\end{equation}
where $h_k^s$ is a normalized output of the target layer $h$ for the sample $k$:
\begin{equation}
    h_k^s = \text{softmax}(h_k)
\end{equation}
$y_k$ is its the ground truth label, $N$ is the number of samples in the batch, $N_s$ is number of samples that have the same label, and $f_l(h_i,h_j)$ is the function defined in~\autoref{eq:f_l}.
Note that this loss component $\mathcal{L}_{\text{multi}}$ works on the outputs of the target layer, and therefore, it affects the whole network behind the layer on which it is applied, overcoming the local properties of the $\mathcal{L}_{\text{single}}$ loss.

\subsection{Unsupervised learning}
Although our main focus in the presented experiments is on a binary classification task, both of our proposed loss components can be used in unsupervised learning as well.
The loss component $\mathcal{L}_{\text{single}}$ does not require any labels so it can be used without modifications. The loss component $\mathcal{L}_{\text{multi}}$ can be applied to unlabeled data by just taking the summations without consideration of labels of the samples as follows:
\begin{equation}
    \label{eq:loss_multi_unlabeled_sum}
    \mathcal{L}_{\text{multi}_2} = \frac{1}{N^2} \sum_{i=1}^{N}\sum_{j=1}^{N} 
    f_l(h_i^s,h_j^s) + f_l(h_j^s,h_i^s)
\end{equation}

For example, as autoencoder models is a common choice to learn representations to use in a downstream task, the proposed loss components can be easily applied to its cost function as follows:
\begin{equation}
    \label{eq:autoencoder_loss}
    \mathcal{L}_{ae} = (1 - \alpha) * \frac{1}{N} \sum_{i=1}^{N} ||X_i-\hat{X_i}||^2 + \alpha * \mathcal{L}_{\text{multi}}
\end{equation}
where the first part is a standard reconstruction cost for autoencoder, the second is the proposed loss component, and $\alpha$ is a hyperparameter reflecting how much importance is given to it.

\begin{table}[h]
\caption{Adjusted Mutual Information (AMI) and Normalized Mutual Information (NMI) scores for the MNIST strokes sequences experiments}
\label{tbl:experiments_mnist_results}
\vskip 0.15in
\begin{center}
\begin{small}
\begin{sc}
\begin{tabular}{lcc}
\hline
\abovespace\belowspace
Model                                       & AMI               & NMI               \\
\hline
\abovespace
Baseline                                    & 0.467             & 0.477             \\
Baseline + \verb|DeCov|                     & 0.287             & 0.313             \\
Baseline + \verb|XCov|                      & 0.525             & 0.547             \\
Baseline + $\mathcal{L}_{\text{single}}$    & \textbf{0.544}    & \textbf{0.553}    \\
\belowspace
Baseline + $\mathcal{L}_{\text{multi}}$     & 0.502             & 0.523             \\ 
\hline
\end{tabular}
\end{sc}
\end{small}
\end{center}
\end{table}

\begin{table*}[t]
\caption{Adjusted Mutual Information (AMI) and Normalized Mutual Information (NMI) scores for the CIFAR-10 experiments}
\label{tbl:experiments_cifar10_results}
\vskip 0.15in
\begin{center}
\begin{small}
\begin{sc}

\begin{tabular}{lcccc}
\hline
\abovespace
\multirow{2}{*}{Model} & \multicolumn{2}{l}{Validation set} & \multicolumn{2}{l}{Test set}    \\
\belowspace
                       & AMI              & NMI             & AMI            & NMI            \\
\hline
\abovespace
Baseline               & 0.198            & 0.204           & 0.194          & 0.199          \\
Baseline + \verb|DeCov|       & 0.175            & 0.191           & 0.176          & 0.191          \\
Baseline + \verb|XCov|        & 0.320             & 0.327           & 0.321          & 0.326          \\
Baseline + $\mathcal{L}_{\text{single}}$   & 0.238            & 0.245           & 0.239          & 0.246          \\
\belowspace
Baseline + $\mathcal{L}_{\text{multi}}$     & \textbf{0.376}   & \textbf{0.384}  & \textbf{0.376} & \textbf{0.385}\\
\hline
\end{tabular}

\end{sc}
\end{small}
\end{center}
\end{table*}

\subsection{Hyperparameter $m$}
One important choice to be made while using the proposed loss components is the value of the hyperparameter $m$. A larger value of $m$ corresponds to a larger margin between the rows of the weights matrix in case of $\mathcal{L}_{\text{single}}$ and a larger margin between the activations of the target layer in case of $\mathcal{L}_{\text{multi}}$. The smaller the value of $m$, the less influence the proposed loss components have. 

In our experiments, we found that the proposed loss component $\mathcal{L}_{\text{single}}$ is relatively stable with respect to the choice of $m$, and generally performs better with larger values (in the range 5-10). In case of the loss component $\mathcal{L}_{\text{multi}}$, we found that even a small value of the margin $m$ (0.1 - 1) disentangles the learned representations better and consequently leads to substantial improvements in the AMI score.

In all of the reported experiments, we found that the proposed loss component with a reasonably chosen $m$ does not hurt the model's performance in the classification task.

\begin{figure*}[th!]
\vskip 0.2in
\begin{center}

\begin{subfigure}{1\columnwidth}
\centering
  \includegraphics[width=0.8\columnwidth,clip, trim=0.1cm 1.25cm 0.1cm 0.1cm]{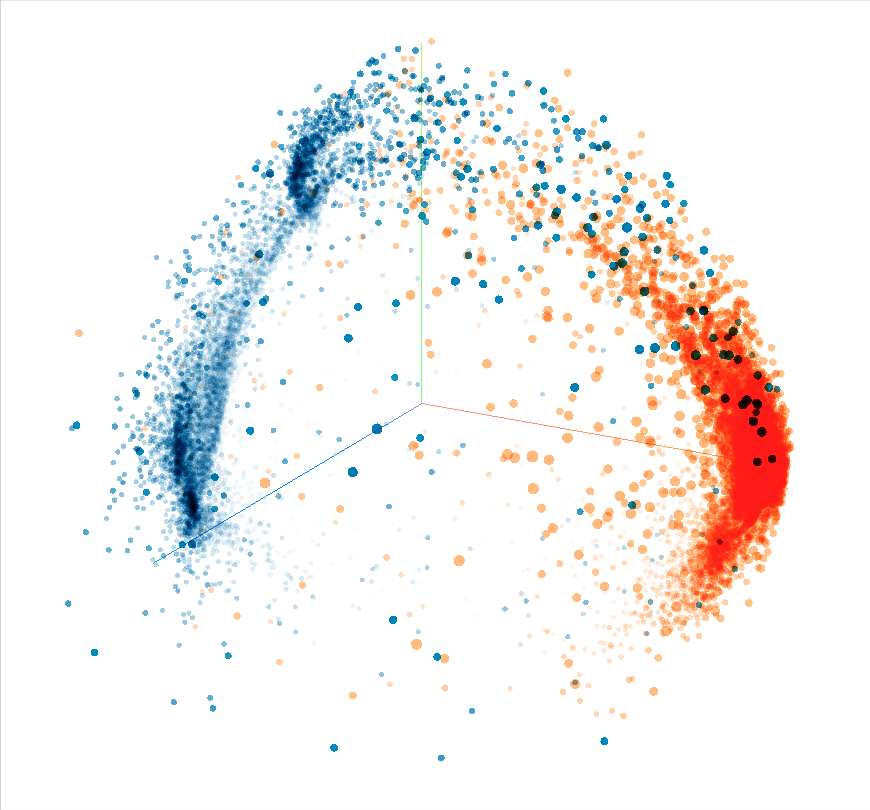}
  \caption{Without the proposed loss component, colored by binary labels}
  \label{fig:mnist_pca_groups_without_loss}
\end{subfigure}
\begin{subfigure}{1\columnwidth}
\centering
  \includegraphics[width=0.8\columnwidth,clip, trim=0.1cm 0.1cm 0.1cm 0.1cm]{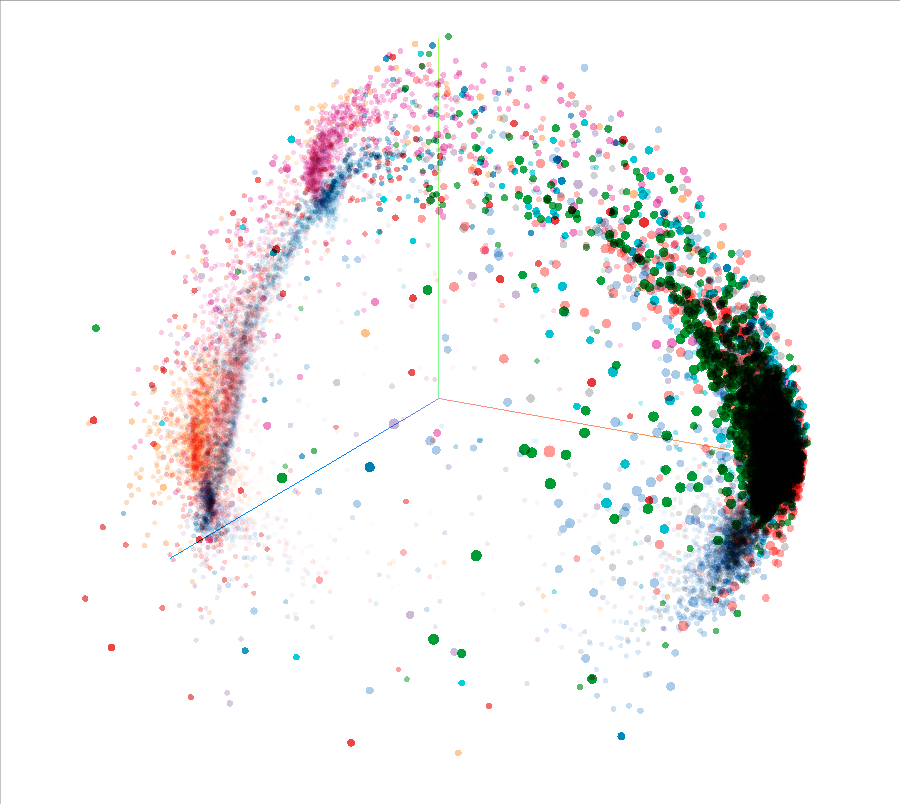}
  \caption{Without the proposed loss component, colored by classes}
  \label{fig:mnist_pca_classes_without_loss}
\end{subfigure}

\begin{subfigure}{1\columnwidth}
\centering
  \includegraphics[width=0.8\columnwidth,clip, trim=0.1cm 0.1cm 0.1cm 0.1cm]{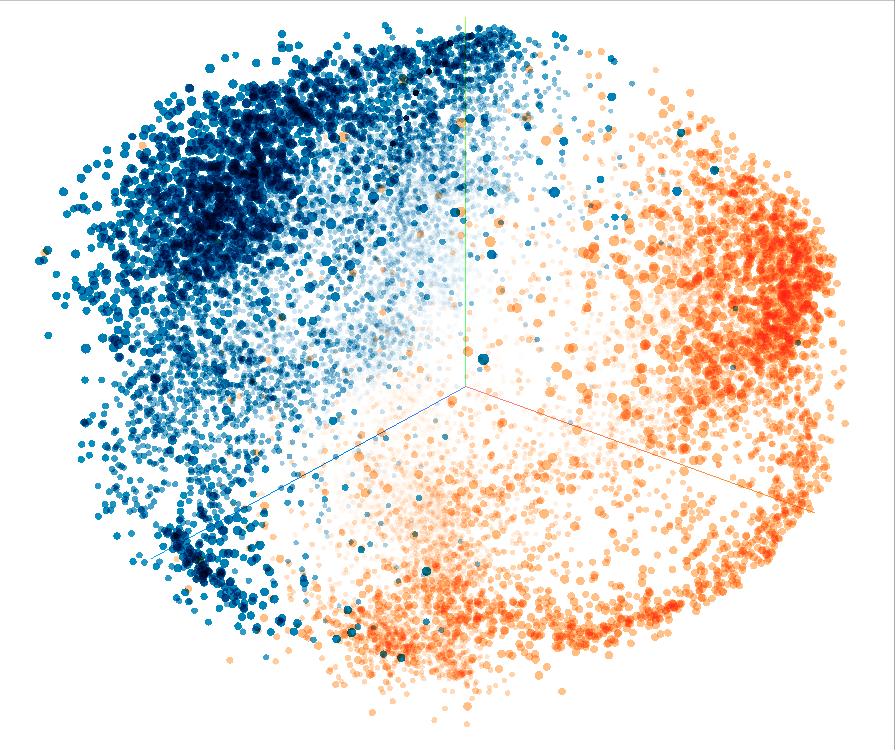}
  \caption{With the proposed loss component $\mathcal{L}_{\text{multi}}$, colored by binary labels}
  \label{fig:mnist_pca_groups_with_loss}
\end{subfigure}
\begin{subfigure}{1\columnwidth}
\centering
  \includegraphics[width=0.8\columnwidth,clip, trim=0.1cm 0.1cm 0.1cm 0.1cm]{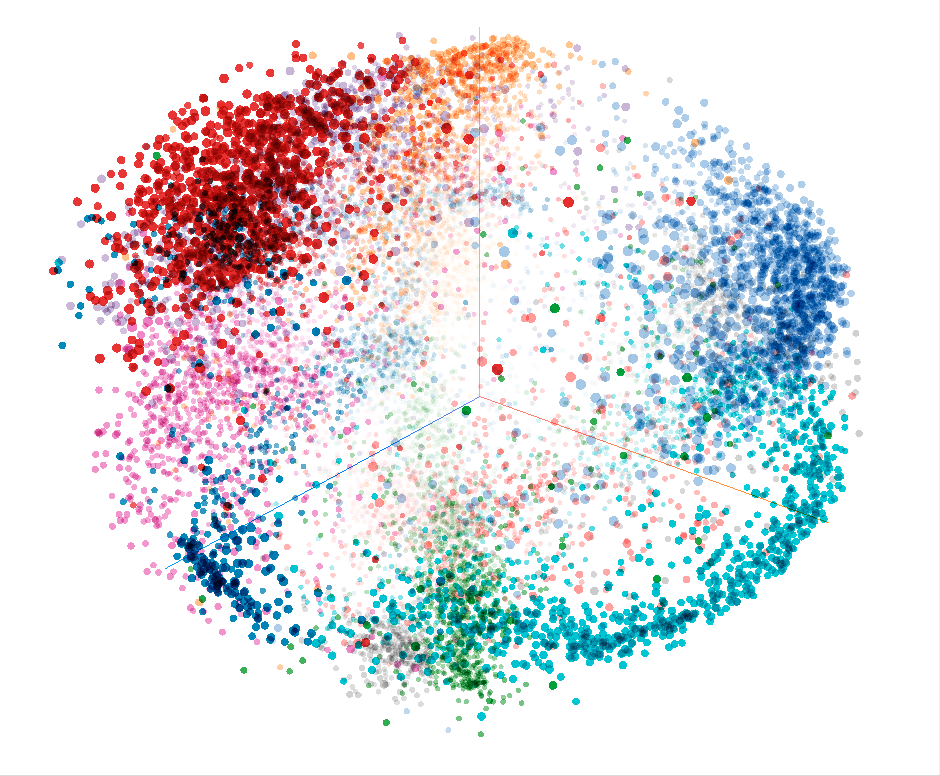}
  \caption{With the proposed loss component $\mathcal{L}_{\text{multi}}$, colored by classes}
  \label{fig:mnist_pca_classes_with_loss}
\end{subfigure}

\caption{PCA visualizations of the learned representations on the MNIST strokes sequences dataset. See the~\autoref{sec:discussion} for details.}
\label{fig:mnist_pca}
\end{center}
\vskip -0.2in
\end{figure*}

\section{Experiments}
\label{sec:experiments}

We performed experiments on the MNIST strokes sequences dataset~\cite{de2016incremental}\footnote{\url{https://github.com/edwin-de-jong/mnist-digits-stroke-sequence-data}} to validate our hypotheses in case of an RNN model. This  dataset contains pen strokes, automatically generated from the original MNIST dataset~\cite{lecun1998mnist}. Although the generated sequences do not always reflect a choice a human would made in order to write a digit, the strokes are consistent across the dataset.

To validate our hypothesis on a more complex dataset, we also experimented with the CIFAR-10 dataset~\cite{krizhevsky2009learning} and a more complex CNN model based on the VGG-16 architecture \cite{simonyan2014very}.

Our experiments therefore cover two types of neural networks most commonly used in modern research: Recurrent Neural Networks and Convolutional Neural Networks, which are used as baseline models in our experiments.

We implemented the models used in all experiments with TensorFlow~\cite{abadi2016tensorflow} and used Adam optimizer~\cite{kingma2014adam} to train the them. Hyperparameters for the baseline models were chosen based on the Adjusted Mutual Information (AMI) score on the validation set and then used in the subsequent models. Hence, the performance of the baseline system is close to an empirical maximum that these models are able to achieve. 

\subsection{MNIST strokes sequences experiments}
\label{sec:experiments_mnist}
For this experiment, we split the examples into two groups: samples belonging to the classes from 0 to 4 were assigned to the first group, and samples belonging to the classes from 5 to 9 were assigned to the second group. The model is trained to predict the \textit{group} of a given sample and does not have any access to the underlying classes. This experiment is a simplified version of the example we discussed in the introduction where we wanted to cluster the patients into meaningful groups, while having only binary mortality outcomes, rather than the more fine-grained labels. 

We used the model depicted in the~\autoref{fig:rnn_model} for this experiment. After the models were trained on the binary classification task, we used the output of the penultimate layer $\text{FC}_2$ to perform KMeans clustering and evaluated the quality of the produced clustering using the original class labels as ground truth assignments.

We compared our loss components $\mathcal{L}_{\text{single}}$ and $\mathcal{L}_{\text{multi}}$ with the \verb|DeCov| regularizer~\citep{cogswell2015reducing} and \verb|XCov| penalty~\citep{cheung2014discovering} as these losses use similar ideas, even though they do not directly target the task of improving the quality of clustering. We did not do comparison with the work of~\citet{liao2016learning} and~\citet{xie2016unsupervised} as they belong to the second group of methods which requires a more complicated training procedure, whereas the loss components proposed here are simple to apply to an arbitrary cost function and do not require any changes in the training procedure.

We report the average of the Adjusted Mutual Information (AMI) and Normalized Mutual Information (NMI) scores~\citep{vinh2010information} across three runs in the~\autoref{tbl:experiments_mnist_results}.

\subsection{CIFAR-10 experiments}
\label{sec:experiments_cifar10}
As in the MNIST strokes sequences experiments, we split the examples in two groups: samples belonging to the classes ``airplan'', ``automobile'', ``bird'', ``cat'', and ``deer'' were assigned to the first group, and samples belonging to the classes ``dog'', ``frog'', ``horse'', ``ship'', ``truck'' were assigned to the second group. Note that this assignment is quite arbitrary as it simply reflects the order of the labels of the classes in the dataset (namely, the labels 0-4 for the first group and the labels 4-9 for the second group). All groups contain rather different types of objects, both natural and human-made.

For the experiments on the CIFAR-10 dataset, we used a CNN model based on the VGG-16 architecture, depicted on the~\autoref{fig:cnn_model}. We discarded the bottom fully connected and convolutional layers as, perhaps, they are too big for this dataset. Instead, we appended three convolutional layers to the output of \verb|pool3| layer with number of filters 256, 128 and 8 correspondingly. The first two layers use 3x3 convolutions, and the last layer uses 1x1 convolutions. After that, we pass the output trough a fully-connected layer of size 15 ($\text{FC}_1$), which will be producing the representations to be used in clustering, and a fully connected layer of size 1 ($\text{FC}_2$) with the sigmoid activation function to produce a binary classification decision.

As in the~\autoref{sec:experiments_mnist}, we compare the proposed losses with \verb|DeCov| and \verb|XCov| and report the AMI and NMI scores in~\autoref{tbl:experiments_cifar10_results}.

\section{Implementation details}
\label{sec:implementation}

Despite the fact the the proposed loss components can be directly implemented using two nested \texttt{for} loops, such implementation will not be computationally efficient, as it will lead to a big computational graph operating on separate vectors without using full advantages of highly optimized parallel matrix computations on GPU. Therefore, it is desirable to have an efficient implementation that can use full advantage of modern GPUs. We have developed such an efficient implementation that significantly accelerates the computation of the loss component in return for a higher memory consumption by creating two matrices that contain all combinations of $d_i$ and $d_j$ from the summations in the~\autoref{eq:loss_single_sum} and performing the operations to calculate the loss on them.
We have made our implementation for TensorFlow~\cite{abadi2016tensorflow} publicly available on GitHub\footnote{\url{http://github.com/placeholder/}} alongside with aforementioned models from the~\autoref{sec:experiments_mnist} and the~\autoref{sec:experiments_cifar10}.

It is worth noting that since the loss component $\mathcal{L}_{\text{single}}$ operates directly on the weights of the target layer, its computational complexity does not depend on the size of the batch. Instead, it depends on the size of that layer. In contrast, the $\mathcal{L}_{\text{multi}}$ operates on the activations of the target layer on all samples in the batch, and its computational complexity depends on the number of samples in the batch. In practice, using the implementation described above, we were able to train models with batch size of 512 and higher without exhausting the GPU's memory.

\section{Discussion}
\label{sec:discussion}

As we can see from \autoref{fig:neurons} and \autoref{fig:neurons_cifar10}, during the binary classification task on both datasets without the proposed loss component the models tend to learn representations that is specific to the target binary label, even though the samples within one group come from different classes. The model learns to use mostly just two neurons to discriminate between the target groups and hardly uses the rest of the neurons in the layer.
We observe this behaviour across different types of models and datasets: an RNN model applied to a timeseries dataset and an CNN model applied to an image classification dataset behave in the exactly the same way. 
Both proposed loss components $\mathcal{L}_{\text{single}}$ and $\mathcal{L}_{\text{multi}}$ force the model to produce disentangled representations, and we can see how it changes the patterns of activations in the target layer. It is easy to see in \autoref{fig:neurons_with_loss} that the patterns of activations learned by the networks roughly correspond to underlying classes, despite the fact that the network did not have access to them during the training. This pattern is not as easy to see in case of CIFAR-10 dataset (see the~\autoref{fig:neurons_cifar10_with_loss}), but we can observe that the proposed loss component nevertheless forced the network to activate different neurons for different classes, leading to a better AMI score on the clustering task.

In order to further investigate the representations learned by the model, we visualized the representations of samples from the MNIST strokes sequences dataset in \autoref{fig:mnist_pca} using TensorBoard. \autoref{fig:mnist_pca_groups_without_loss} and \autoref{fig:mnist_pca_classes_without_loss} in the top row depict the representations learned by the baseline model, colored according to the binary label and the underlying classes, respectively. \autoref{fig:mnist_pca_groups_with_loss} and \autoref{fig:mnist_pca_classes_with_loss} in the bottom row depict the  representations of the same samples, learned by the model with the $\mathcal{L}_{\text{multi}}$ loss component, colored in the same way. It is easy to see that the $\mathcal{L}_{\text{multi}}$ indeed forced the model to learn disentangled representations of the input data. Note how the baseline model learned dense clusters of objects, with samples from the same group (but different classes) compactly packed in the same area. In contrast, the model with the proposed loss component learned considerably better representations which disentangle samples belonging to different classes and placed the them more uniformly in the space.

\begin{figure}[ht]
\vskip 0.2in
\begin{center}
\centering
    \includegraphics[width=0.8\columnwidth]{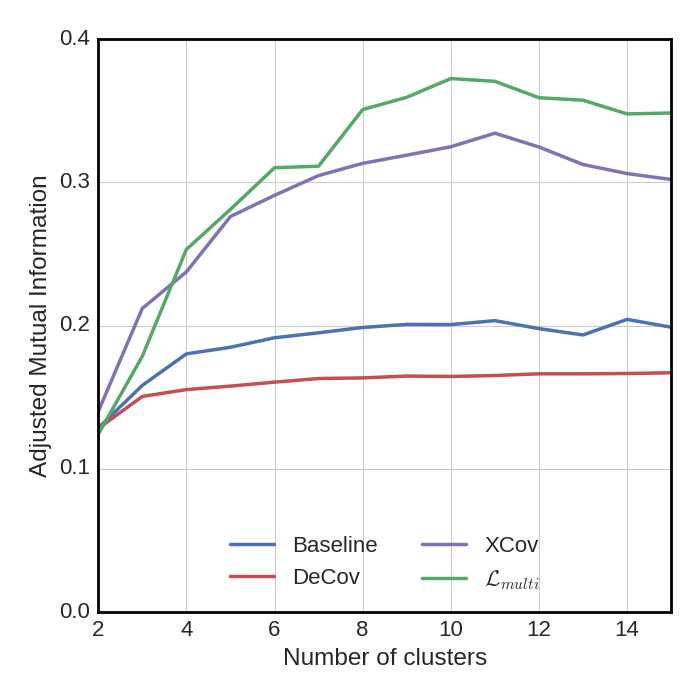}
\caption{Number of clusters and the corresponding AMI score on the CIFAR-10 dataset}
\label{fig:nb_clusters_methods}
\end{center}
\vskip -0.2in
\end{figure}

In the real world, the number of clusters is rarely known beforehand. To systematically examine the stability of the proposed loss component, we plotted the Adjusted Mutual Information scores for the baselines methods and $\mathcal{L}_{\text{multi}}$ loss component with respect to the number of clusters in \autoref{fig:nb_clusters_methods}, using the CIFAR-10 dataset. As can be seen from \autoref{fig:nb_clusters_methods}, our loss component consistently outperforms the previously proposed methods regardless the number of clusters.

\section{Conclusion}
\label{sec:conclusion}

In this paper, we proposed two novel loss components that substantially improve the quality of KMeans clustering using the representations of the input data learned by the model. We performed a comprehensive set of experiments using two important model types (RNNs and CNNs) and different datasets, demonstrating that the proposed loss components consistently increase the Adjusted Mutual Information score by a significant margin and outperform previously proposed methods. In addition, we analyzed the representations learned by the network by visualizing the activation patterns and relative positions of the samples in the learned space, and the visualizations show that the proposed loss components indeed force the network to learn disentangled representations.


 

\clearpage

\bibliography{related_work}

\begin{thebibliography}{17}
\providecommand{\natexlab}[1]{#1}
\providecommand{\url}[1]{\texttt{#1}}
\expandafter\ifx\csname urlstyle\endcsname\relax
  \providecommand{\doi}[1]{doi: #1}\else
  \providecommand{\doi}{doi: \begingroup \urlstyle{rm}\Url}\fi

\bibitem[Abadi et~al.(2016)Abadi, Agarwal, Barham, Brevdo, Chen, Citro,
  Corrado, Davis, Dean, Devin, et~al.]{abadi2016tensorflow}
Abadi, Mart{\'\i}n, Agarwal, Ashish, Barham, Paul, Brevdo, Eugene, Chen,
  Zhifeng, Citro, Craig, Corrado, Greg~S, Davis, Andy, Dean, Jeffrey, Devin,
  Matthieu, et~al.
\newblock Tensorflow: Large-scale machine learning on heterogeneous distributed
  systems.
\newblock \emph{arXiv preprint arXiv:1603.04467}, 2016.

\bibitem[Cheung et~al.(2014)Cheung, Livezey, Bansal, and
  Olshausen]{cheung2014discovering}
Cheung, Brian, Livezey, Jesse~A, Bansal, Arjun~K, and Olshausen, Bruno~A.
\newblock Discovering hidden factors of variation in deep networks.
\newblock \emph{arXiv preprint arXiv:1412.6583}, 2014.

\bibitem[Cogswell et~al.(2015)Cogswell, Ahmed, Girshick, Zitnick, and
  Batra]{cogswell2015reducing}
Cogswell, Michael, Ahmed, Faruk, Girshick, Ross, Zitnick, Larry, and Batra,
  Dhruv.
\newblock Reducing overfitting in deep networks by decorrelating
  representations.
\newblock \emph{arXiv preprint arXiv:1511.06068}, 2015.

\bibitem[de~Jong(2016)]{de2016incremental}
de~Jong, Edwin~D.
\newblock Incremental sequence learning.
\newblock \emph{arXiv preprint arXiv:1611.03068}, 2016.

\bibitem[Karpathy et~al.(2015)Karpathy, Johnson, and
  Fei-Fei]{karpathy2015visualizing}
Karpathy, Andrej, Johnson, Justin, and Fei-Fei, Li.
\newblock Visualizing and understanding recurrent networks.
\newblock \emph{arXiv preprint arXiv:1506.02078}, 2015.

\bibitem[Kingma \& Ba(2014)Kingma and Ba]{kingma2014adam}
Kingma, Diederik and Ba, Jimmy.
\newblock Adam: A method for stochastic optimization.
\newblock \emph{arXiv preprint arXiv:1412.6980}, 2014.

\bibitem[Krizhevsky \& Hinton(2009)Krizhevsky and
  Hinton]{krizhevsky2009learning}
Krizhevsky, Alex and Hinton, Geoffrey.
\newblock Learning multiple layers of features from tiny images.
\newblock 2009.

\bibitem[Krizhevsky et~al.(2012)Krizhevsky, Sutskever, and
  Hinton]{krizhevsky2012imagenet}
Krizhevsky, Alex, Sutskever, Ilya, and Hinton, Geoffrey~E.
\newblock Imagenet classification with deep convolutional neural networks.
\newblock In \emph{Advances in neural information processing systems}, pp.\
  1097--1105, 2012.

\bibitem[LeCun et~al.(1998)LeCun, Cortes, and Burges]{lecun1998mnist}
LeCun, Yann, Cortes, Corinna, and Burges, Christopher~JC.
\newblock The mnist database of handwritten digits, 1998.

\bibitem[Li et~al.(2016)Li, Chen, Hovy, and Jurafsky]{li2016visualizing}
Li, Jiwei, Chen, Xinlei, Hovy, Eduard, and Jurafsky, Dan.
\newblock Visualizing and understanding neural models in nlp.
\newblock In \emph{Proceedings of NAACL-HLT}, pp.\  681--691, 2016.

\bibitem[Liao et~al.(2016)Liao, Schwing, Zemel, and Urtasun]{liao2016learning}
Liao, Renjie, Schwing, Alex, Zemel, Richard, and Urtasun, Raquel.
\newblock Learning deep parsimonious representations.
\newblock In \emph{Advances in Neural Information Processing Systems}, pp.\
  5076--5084, 2016.

\bibitem[Russakovsky et~al.(2015)Russakovsky, Deng, Su, Krause, Satheesh, Ma,
  Huang, Karpathy, Khosla, Bernstein, Berg, and Fei-Fei]{ILSVRC15}
Russakovsky, Olga, Deng, Jia, Su, Hao, Krause, Jonathan, Satheesh, Sanjeev, Ma,
  Sean, Huang, Zhiheng, Karpathy, Andrej, Khosla, Aditya, Bernstein, Michael,
  Berg, Alexander~C., and Fei-Fei, Li.
\newblock {ImageNet Large Scale Visual Recognition Challenge}.
\newblock \emph{International Journal of Computer Vision (IJCV)}, 115\penalty0
  (3):\penalty0 211--252, 2015.
\newblock \doi{10.1007/s11263-015-0816-y}.

\bibitem[Simonyan \& Zisserman(2014)Simonyan and Zisserman]{simonyan2014very}
Simonyan, Karen and Zisserman, Andrew.
\newblock Very deep convolutional networks for large-scale image recognition.
\newblock \emph{arXiv preprint arXiv:1409.1556}, 2014.

\bibitem[Susskind et~al.(2010)Susskind, Anderson, and
  Hinton]{susskind2010toronto}
Susskind, Joshua, Anderson, Adam, and Hinton, Geoffrey~E.
\newblock The toronto face dataset.
\newblock \emph{U. Toronto, Tech. Rep. UTML TR}, 1:\penalty0 2010, 2010.

\bibitem[Vinh et~al.(2010)Vinh, Epps, and Bailey]{vinh2010information}
Vinh, Nguyen~Xuan, Epps, Julien, and Bailey, James.
\newblock Information theoretic measures for clusterings comparison: Variants,
  properties, normalization and correction for chance.
\newblock \emph{Journal of Machine Learning Research}, 11\penalty0
  (Oct):\penalty0 2837--2854, 2010.

\bibitem[Xie et~al.(2016)Xie, Girshick, and Farhadi]{xie2016unsupervised}
Xie, Junyuan, Girshick, Ross, and Farhadi, Ali.
\newblock Unsupervised deep embedding for clustering analysis.
\newblock In \emph{International Conference on Machine Learning (ICML)}, 2016.

\bibitem[Zeiler \& Fergus(2014)Zeiler and Fergus]{zeiler2014visualizing}
Zeiler, Matthew~D and Fergus, Rob.
\newblock Visualizing and understanding convolutional networks.
\newblock In \emph{European conference on computer vision}, pp.\  818--833.
  Springer, 2014.

\end{thebibliography}
\bibliographystyle{icml2017}

\end{document}